\documentclass[runningheads]{llncs}
\usepackage{graphicx}
\usepackage{comment}
\usepackage{amsmath,amssymb} 
\usepackage{color}
\usepackage{wrapfig}
\usepackage{floatflt}
\usepackage{subfig}
\newcommand\norm[1]{\left\lVert#1\right\rVert}

\begin{document}
\pagestyle{headings}
\mainmatter
\def\ECCVSubNumber{6059}  

\title{High Resolution Zero-Shot Domain Adaptation of Synthetically Rendered Face Images} 

\titlerunning{Zero-Shot Synthetic $\rightarrow$ Real Transfer}
%
\author{Stephan J. Garbin \and
Marek Kowalski \and
Matthew Johnson \and
Jamie Shotton}
\authorrunning{S. J. Garbin et al.}
%
\institute{Microsoft}
\maketitle

\begin{abstract}
  Generating photorealistic images of human faces at scale remains a prohibitively difficult task using computer graphics approaches. This is because these require the simulation of light to be photorealistic, which in turn requires physically accurate modelling of geometry, materials, and light sources, for both the head and the surrounding scene. Non-photorealistic renders however are increasingly easy to produce. In contrast to computer graphics approaches, generative models learned from more readily available 2D image data have been shown to produce samples of human faces that are hard to distinguish from real data. The process of learning usually corresponds to a loss of control over the shape and appearance of the generated images. For instance, even simple disentangling tasks such as modifying the hair independently of the face, which is trivial to accomplish in a computer graphics approach, remains an open research question. In this work, we propose an algorithm that matches a non-photorealistic, synthetically generated image to a latent vector of a pretrained StyleGAN2 model which, in turn, maps the vector to a photorealistic image of a person of the same pose, expression, hair, and lighting. In contrast to most previous work, we require no synthetic training data. To the best of our knowledge, this is the first algorithm of its kind to work at a resolution of 1K and represents a significant leap forward in visual realism.
\end{abstract}

\section{Introduction}

\begin{figure}[htp]
  \centering
  \includegraphics[width=\textwidth]{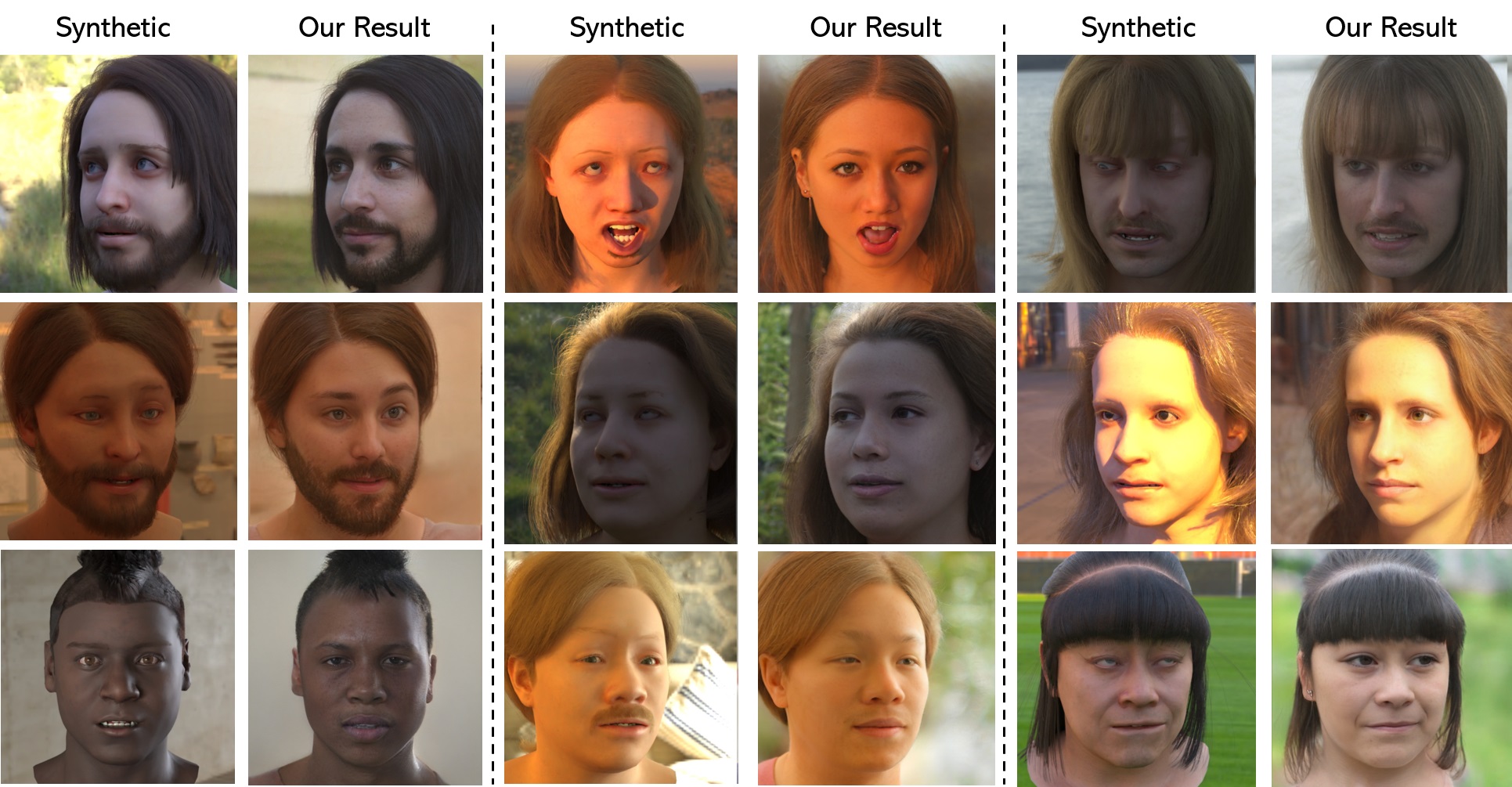}
  \caption{Pairs of synthetic input images, and samples from our algorithm at 1K resolution.  Despite the real dataset used (FFHQ) mostly containing images of people looking directly at the camera with neutral or smiling expressions, we are able to match more diverse poses and facial expressions. Best viewed zoomed in, and in colour.}
  \label{fig:teaser}
\end{figure}

Generating photorealistic images of human faces remains a challenge in computer graphics. While we consider the arguably easier problem of still images as opposed to animated ones, we note that both pose unsolved research questions. This is because of the complicated and varied appearance of human tissue found in the skin \cite{wrenninge2017path}, eyes \cite{10.1145/2897824.2925962} and teeth of the face region. The problem is further complicated by the fact that humans are highly attuned to the appearance of faces and thus skilled at spotting any unnatural aspect of a synthetic render \cite{uncannyValleyArticle}.
Machine learning has recently seen great success in generating still images of faces that are nearly indistinguishable from the domain of natural images to a non-expert observer. This gives methods like StyleGAN2 (SG2) \cite{karras2019analyzing} a clear advantage over computer graphics if the goal is to generate photorealistic image samples only. The limitation of models like SG2 is that we get RGB data only, and that such samples are often only useful if annotations such as head pose, UVs, or expression parameters are available for downstream tasks. The second major issue is that generative models necessarily inherit the bias of the data they were trained on. For large image collections, this may be hard to assess \cite{DBLP:journals/corr/abs-1811-03259}. In computer graphics on the other hand, annotations such as UVs can be trivially obtained for an image. Since the assets that define the data input into the renderer need to be explicitly created, bias control becomes more feasible.

In this paper, we propose to play to the strengths of both fields by using machine learning to change the appearance of non-photorealistic renders to be more natural, while keeping semantics such as the shape, the expression of the face, and the lighting consistent. This means that the annotations obtained from the renders are still valid for the images after domain transfer. Because we do not require photo-realism from the synthetic renders, we can produce them at scale and with significant variety using a traditional graphics pipeline. 

In contrast to much recent work in the field, we require no synthetic images for training at all, and thus no paired data. In fact, our method only requires a pre-trained StyleGAN2 model, and a small number of manual annotations from the data it was trained on as outlined below. For a given synthetic image, our methods works best if masks of the hair and background are available. These can be easily obtained from any renderer.
Our method works by finding an embedding in the latent space of SG2 that produces an image which is perceptually similar to a synthetic sample, but still has the characteristic features of the data the GAN was trained with. In terms of the scale space of an image \cite{BURT1987671,BURT198120}, we attempt to match the coarser levels of an image pyramid to the synthetic data, and replace fine detail with that of a photorealistic image. Another way to interpret this is that we attempt to steer StyleGAN2 with the help of synthetic data \cite{DBLP:journals/corr/abs-1907-07171}.

While embedding images in the latent space of SG2 is not a new concept \cite{DBLP:journals/corr/abs-1904-03189,abdal2019image2stylegan}, the issue with using existing approaches is that they either do not, or struggle to, enforce constraints to keep the results belonging to the distribution of real images. In fact, the authors in \cite{DBLP:journals/corr/abs-1904-03189} explicitly note that almost any image can be embedded in a StyleGAN latent space. If closeness to the domain of real images is not enforced, we simply get an image back from the generator that looks exactly like the synthetic input whose appearance we wish to change.

We make the observation that samples from the prior distribution usually \emph{approximately} form a convex set, \textit{i.e.} that convex combinations of any set of such points mapped through the generator are statistically similar to samples from the data distribution the GAN was trained on. We also note that showing interpolations between pairs of latent vectors is a common strategy of evaluating the quality of the latent embeddings of generative models \cite{DBLP:journals/corr/RadfordMC15}. As part of our method, we propose an algorithm, Convex Set Approximate Nearest Neighbour Search (CS-ANNS), which can be used to traverse the latent space of a generative model while ensuring that the reconstructed images closely adhere to the prior. This algorithm optimises for the combination of a set of samples from the SG2 prior by gradient descent, and is detailed in the method section below.

In summary, our contributions are:
\begin{enumerate}
  \item The first zero-shot domain transfer method to work at 1K and with only limited annotations of the real data, and	
  \item a novel algorithm for approximate nearest neighbour search in the latent spaces of generative models.
\end{enumerate}

\section{Related Work}
\subsection{Generative Models}
Generative models are of paramount importance to deep learning research. In this work, we care about those that map samples from a latent space to images. While many models have been proposed (such as Optimized Maximum Mean Discrepancy \cite{2016arXiv161104488S}, Noise Contrastive Estimation \cite{pmlr-v9-gutmann10a}, Mixture Density Networks \cite{bishop1994mixture}, Neural Autoregressive Distribution Estimators \cite{larochelle2011}, Real-Valued Neural Autoregressive Distribution Estimators \cite{Uria2013}, Diffusion Process Models \cite{DBLP:journals/corr/Sohl-DicksteinW15}), the most popular ones are the family of Variational Autoencoders (VAEs) \cite{2014arXiv1401.4082J,2013arXiv1312.6114K}, and Generative Adversarial Networks (GANs) \cite{2014arXiv1406.2661G}. 

Because GANs are (at the time of writing) capable of achieving the highest quality image samples, we focus on them in this work, and specifically on the current state of the art for face images, StyleGAN2 (SG2) \cite{karras2019analyzing}. In any GAN, a neural sampler called the generator is trained to transform samples from a simple distribution to the true distribution over the data space. A second network, called the discriminator, is trained to differentiate samples produced by the generator and those from the data space.

Our method takes as input only the pretrained \emph{generator} of SG2, and while we backpropagate through it, we do not modify its weights as part of our algorithm. Since SG2 uses a variant of Adaptive Instance Normalisation (AdaIn)\cite{huang2017adain}, its latent space is mapped directly to the AdaIn parameters at 18 different layers. We do not use the additional noise inputs at each layer. This way of controlling the generator output via the AdaIn inputs is the same methodology as used in the Image2StyleGAN work \cite{DBLP:journals/corr/abs-1904-03189}. The authors in \cite{DBLP:journals/corr/abs-1904-03189} also consider style transfer by blending two latent codes together, but choose very different image modalities such as cartoons and photographs. We build on their work by defining a process that finds a close nearest neighbour to blend with, thereby creating believable appearance transfer that preserves semantics.

\subsection{Zero-Shot Domain Transfer}
To the best of our knowledge, there are no zero-shot image domain transfer methods in the literature that require only one source domain. By domain adaptation we mean the ability to make images from dataset $A$ look like images from dataset $B$, while preserving content. While one-shot methods like \cite{DBLP:journals/corr/abs-1905-04729} or \cite{DBLP:journals/corr/abs-1806-06029} have been proposed, they work at significantly lower resolution than ours and still require one sample from the target domain.

ZstGAN~\cite{DBLP:journals/corr/abs-1906-00184}, the closest neighbour, requires many source domains (that could for example be extracted from image labels of one dataset). The highest resolution handled in that work is $128^2$, which is signifantly lower than our method. The categories are used to bootstrap the appearance transfer problem, as if multiple datasets were available. Without labels for dividing the data into categories, we were unable to use it as a baseline.

\subsection{Domain Adaptation}
If paired training data from two domains is available, Pix2Pix \cite{DBLP:journals/corr/IsolaZZE16} and its successors (e.g. Pix2Pix HD \cite{wang2017highresolution}, which changes to architecture to use multiple discriminators at multiple scales to produce high resolution images) can be used effectively for domain adaptation.

CycleGAN does not require paired training data \cite{CycleGAN2017}. This brings it closer to the application we consider. However, it still requires a complete dataset of both image modalities. Many improvements have since been suggeted to improve CycleGAN. HarmonicGAN adds an additional smoothness constraint to reduce artefacts in the outputs \cite{DBLP:journals/corr/abs-1902-09727}, Sem-GAN exploits additional information in the source images \cite{SemGANpaper}, as does \cite{albahar2019guided}, Discriminative Region Proposal Adversarial Networks (DRPAN) \cite{DBLP:journals/corr/abs-1711-09554} add steps to fix errors, Geometry-Consistent GANs (GcGAN) \cite{DBLP:journals/corr/abs-1809-05852} use consistency under simple transformations as an additional constraint. Some methods also model a distribution of over possible outputs, such as MUNIT \cite{Huang_2018_ECCV} or FUNIT \cite{Liu_2019_ICCV}.

However, none of these methods are capable of zero-shot domain adaptation.


\section{Method}
\begin{figure}
  \centering
  \includegraphics[width=\textwidth]{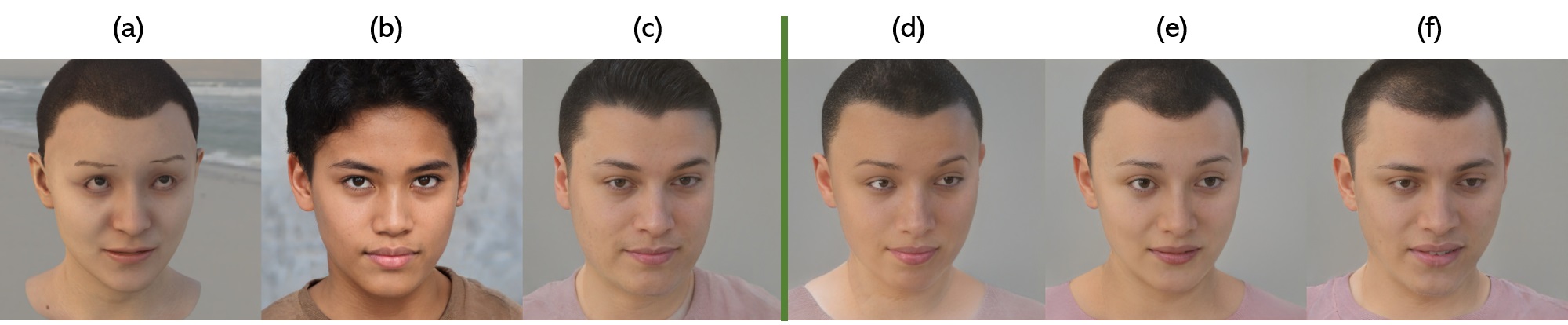}
  \caption{Illustration of the different steps of our method. (a) is the input synthetic render, (b) the output of the sampling in step 1, (c) the result of Convex Set Approximate Nearest Neighbour Search in step 2, and (d-f) results from step 3.}
  \label{fig:method_progression_0}
\end{figure}

In the following, any variable containing $w$ refers to the $18 \times 512$ dimensional inputs of the pretrained SG2 generator, $G$. Any variable prefixed with $I$ refers to an image, either given as input, or obtaining by passing a $w$ through the generator $G$. The proposed method takes as input a synthetically rendered image $I^s$, and returns a series of $w$s that represent domain adapted versions of that input.
\begin{wrapfigure}{R}{0.5\textwidth}
  \centering
  \includegraphics[width=0.5\textwidth]{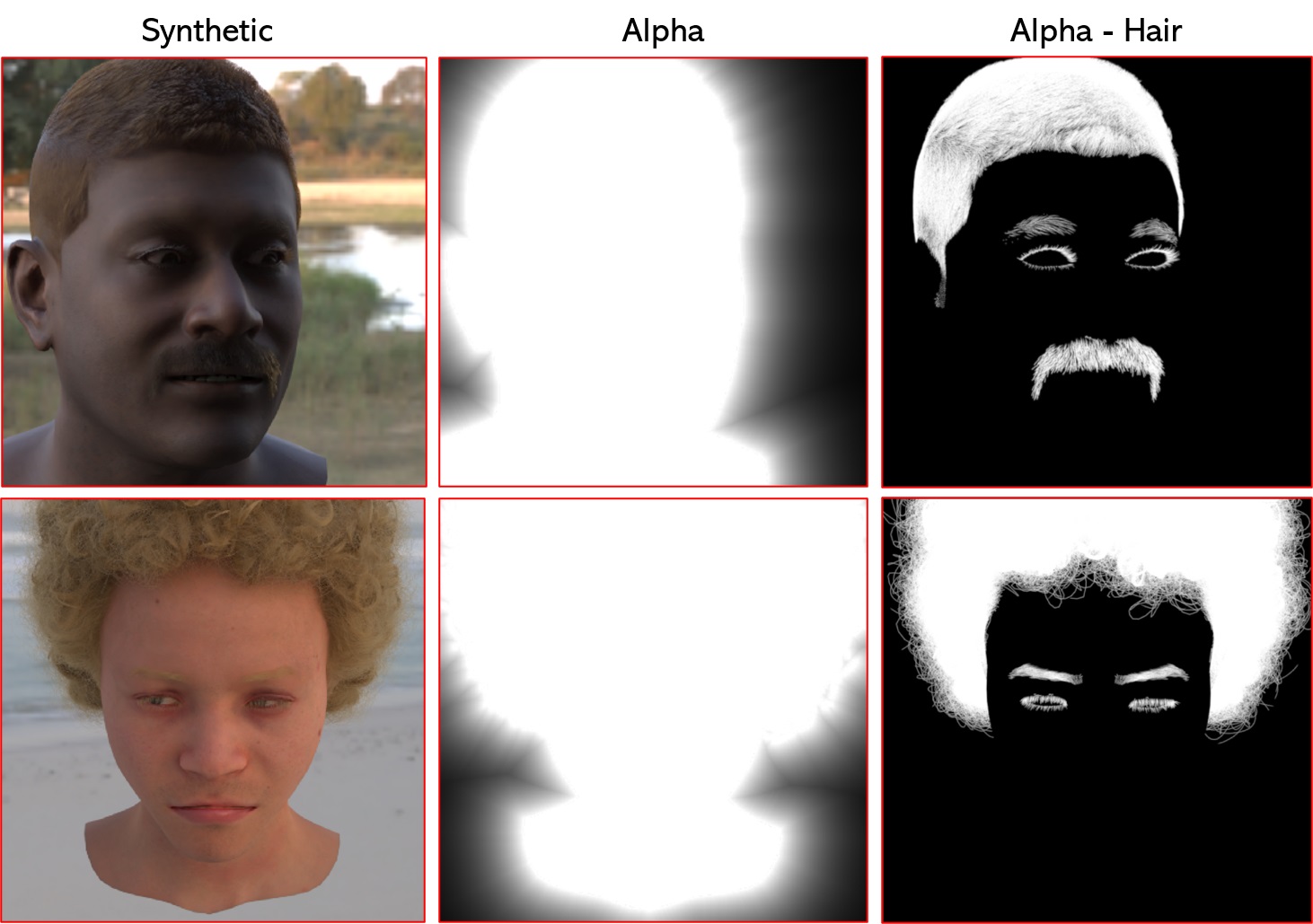}
  \vspace{-\intextsep}
  \caption{Example tuples of renders and alpha masks, $\{I^s, I^a, I^{a_{hair}}\}$, derived from synthetic images. Note that we apply a falloff at sharp boundaries to preserve them as described in the text.}
  \label{fig:alpha}
\end{wrapfigure}

Our algorithm has four stages, each producing results more closely matching the input. In the first, we find the latent code, $w^s$, of an approximate nearest neighbour to a given synthetic input image, $I^s$, by sampling. This is the starting point of our method. For the second step, we propose Convex Set Approximate Nearest Neighbour Search (CS-ANNS), an algorithm to refine the initial sample by traversing the latent space while being strongly constrained to adhere to the prior. This gives us a refined latent code, $w^n$.

In the third step, we fit SG2 to the synthetic image \emph{without any constraint} to obtain another latent code $w^f$ that matches $I^s$ as closely as possible. We can then combine $w^f$ and $w^n$ with varying interpolation weights to obtain a set of final images that strongly resemble $I^s$, but which have the appearance of real photographs. 

Because $w^s$, $w^n$ and the results from step 3 are all valid proposals for the final result, we select the latent code that gives an image as semantically similar to $I^s$ as possible from among them in the fourth and final step. An example of the different steps of our method can be seen in Figure~\ref{fig:method_progression_0}.

We note that the SG2 model used in this section was trained on the FFHQ dataset \cite{DBLP:journals/corr/abs-1812-04948}, a dataset of photographs at high resolution. We use the same preprocessing and face normalisation as the authors of that work.

Since we care about closely matching the face in this work, we construct floating-point alpha masks from the synthetic renders that de-emphasize the background and allow us to separate the hair. This gives us tuples of renders and alpha masks $\{I^s, I^a, I^{a_{hair}}\}$ for each input. We observe that in order to get accurate matching of face boundaries, the sharp opacity edges of $I_a$ that come from the renderer need to be extended outwards from the face. We compute the distance transfer for the face boundary and produce a quickly decaying falloff by mapping the resulting values, remapped to be in the range $0-1$, by $x^{10}$, where $x$ is the output of the distance transform at a pixel. This is illustrated in Figure~\ref{fig:alpha}.

\subsection{Step 1: Sampling}

\begin{wrapfigure}{R}{0.45\textwidth}
  \centering
  \includegraphics[width=0.45\textwidth]{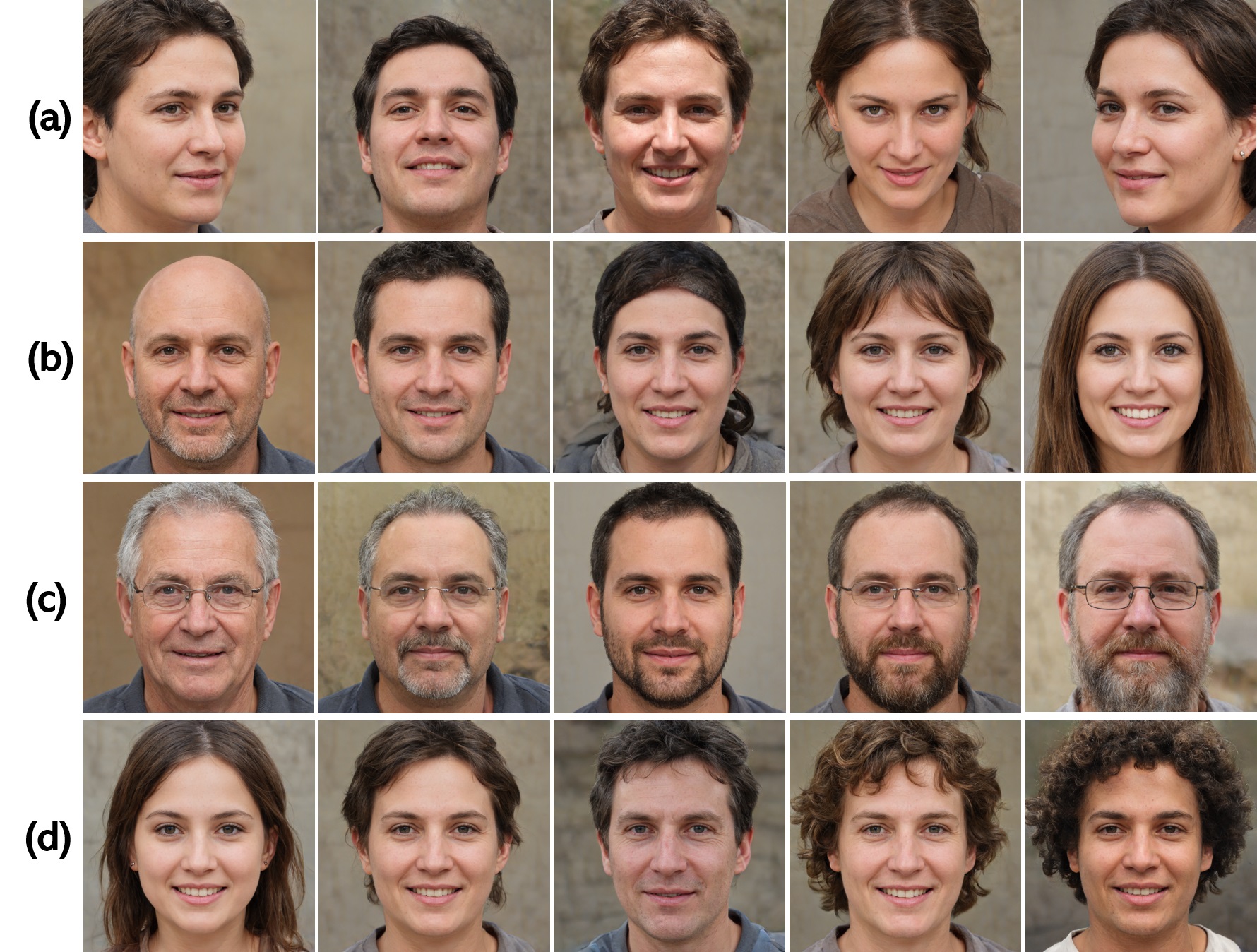}
  \vspace{-\intextsep}
  \caption{Example of adding our control vectors to the 'mean' face of SG2: (a) face angle; (b) hair length (including headgear); (c) beard length; (d) hair curlyness. Note that these can be found by only rough annotations of a small number of samples. Best viewed digitally, and in colour.}
  \label{fig:control_vectors}
  \vspace{-\intextsep}
\end{wrapfigure}

\begin{wrapfigure}{L}{0.5\textwidth}
  \centering
  \includegraphics[width=0.5\textwidth]{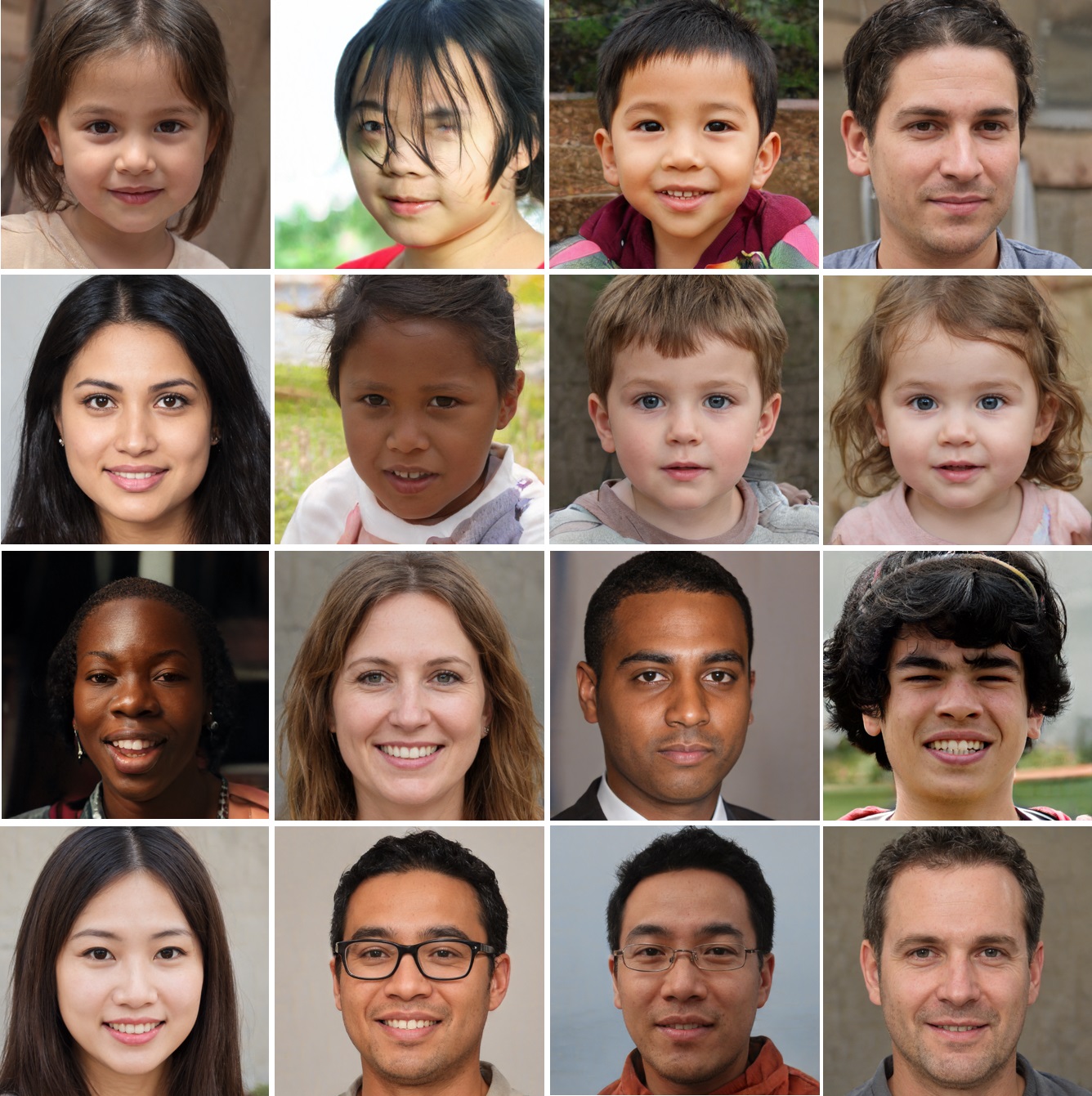}
  \vspace{-\intextsep}
  \caption{Our manually curated set of centroids for sampling.}
  \label{fig:centroids}
  \vspace{-\intextsep}
\end{wrapfigure}

To find a good initialisation, we could sample from the prior of SG2 and take the best match as input to the other steps. However, we found that this could fail to produce convincing results for faces at an angle, under non frontal illumination etc. because our synthetic data is more varied in pose, lighting and ethnicity than FFHQ. To overcome this problem, we annotate a small subset of 2000 samples from SG2 with a series of simple attributes to obtain a set of $33$ control vectors, $v_{control}$. These are detailed in the supplementary material. The effect of adding some of these to the mean face of SG2 is shown in Figure~\ref{fig:control_vectors}. We also select a set of centroids, $v_{centroid}$, to sample around. As can be seen in Figure~\ref{fig:centroids}, these are selected to be somewhat balanced in terms of sex, skin tone and age, and are chosen empirically. We are unable to prove that this leads to greater overall fairness \cite{DBLP:journals/corr/abs-1710-03184}, and acknowledge that this sensitive issue needs closer examination in future work.


The loss used in the sampling step is a combination of the LPIPS distance \cite{DBLP:journals/corr/abs-1801-03924}, an L1 loss with different weights for colour and luminance, and landmark loss based on $68$ points compute with DLIB \cite{6909637}. The loss is computed at a quarter resolution, $256^2$ pixels, after low pass filtering with a Gaussian kernel of size $7$ and standard deviation of $0.15$, and multiplication with the mask $I^a$. We found this crucial to ensure fine detail is retained in the fits. The entire loss function for the sampling step is thus:
\begin{equation}
  \begin{split}
  L_{sampling} = L_{LPIPS}(r(I^s * I^a), r(G(w^s) * I^a))  \\
  + \lambda_{lum} * \norm{y(r(I^s * I^a)) - y(r(G(w^s) * I^a))}^1  \\
  + \lambda_{col} * \norm{u(r(I^s * I^a)) - u(r(G(w^s) * I^a))}^1  \\
  + \lambda_{landm} * \norm{l(r(I^s * I^a)) - l(r(G(w^s) * I^a))}^2,
  \end{split}
  \label{eq:sampling_loss}
\end{equation}
where $r$ is the resampling function that changes image size after Gaussian filtering, $u$ separates out the colour channels in the YUV colour space, $y$ the luminance channel, $G$ is the pretrained SG2 generator, $I^s$ a synthetic image, $w^s$ a latent code sample, and $l$ the landmark detector. $\lambda_{lum}$ is set to $0.1$, $\lambda_{col}$ to $0.01$, and $\lambda_{landm}$ to $1e-5$.

For each sample at this stage of our method, we pick one of the centroids $v_{centroid}$ with uniform probability, and add Gaussian noise to it. We then combine this with a random sample of our control vectors to vary pose, light, expression etc. The i'th sample is thus obtained as:
\begin{equation}
  \begin{split}
  w^{s}_{i} = s(v_{centroid}) + \mathcal{N}(0.0,\sigma^{2}) \\
  + v_{control} * N_{uniform} * 2.0,
\end{split}
\end{equation}
where $s$ is the random centroid selection function, $\sigma^{2} = 0.25$, and $N_{uniform}$ is uniform noise to scale the control vectors.

The output of this stage is simply the best $w^s$ under the loss in Equation~\ref{eq:sampling_loss}, for any of the $512$ samples taken.

\subsection{Step 2: Convex Set Approximate Nearest Neighbour Search}
In step 2, we refine $w^{s}$ while keeping the results constrained to the domain of valid images in SG2. The intuition is that any convex combination of samples from the prior in the latent space also leads to realistic images when decoded through $G$. We highlight that $w^{n}$ is the current point in the latent space, and updated at every iteration. It is initialised with the result from step 1.

At each step, we draw $512$ samples using the same procedure as before, just without adding noise, which now becomes a learnable parameter, $\beta$. At each iteration, each of the $512$ proposal samples $w^{p}$ is obtained as:
\begin{equation}
w^{p}_{i} = s(v_{centroid}) + \mathcal{N}(0.0,\sigma^{2}).
\end{equation}
We blend the centroid candidates $w^{p}_{i}$ using linear interpolation with the current $w^n$ using a uniform random $\alpha$ in the range of $0.25-0.75$, as we found that this stabilised the optimisation. We then optimise for a set of weights, $\alpha$, which determine how the $w^{p}_{i}$s and current $w^n$ are combined. We use sets of $\alpha$ for each of $18$ StyleGAN2 latent space inputs, \textit{i.e.} $\alpha$ is a matrix of shape $[512+1, 18]$ (note how the current $w^n$ is included). We constrain the optimisation to make each row of $\alpha$ sum to $1$ using the softmax function, ensuring a convex combination of the samples. In addition to $\alpha$, we allow learned variation for the control vectors, which are scaled by a learnable parameter $\beta$. Because this last step could potentially lead to solutions far outside the space of plausible images, we clamp $\beta$ to $2.0$. The loss is the same as Equation~\ref{eq:sampling_loss}, just without the landmark term, \textit{i.e.} with $\lambda_{landm}$ set to $0$. 

We use 96 outer iterations for which the $w^{p}$ are redrawn, and $\alpha$ and $\beta$ reset so that the current $w^n$ is the starting point (\textit{i.e.} $\beta$ is set to zero, and $alpha$ to one only for the current $w^n$). For each of these outer loops, we optimise $\alpha$ and $\beta$ using Adam \cite{kingma2014adam} with a learning rate of $0.01$ in an inner loop. We divide the initial learning rate by $10.0$ for every $4$ iterations in that inner loop, and return the best result at any point, which gives us the refined $w^n$. We name this algorithm Convex Set Approximate Nearest Neighbour Search (CS-ANN). More details can be found in the supplementary material.

\subsection{Step 3: Synthetic Fit and Latent Code Interpolation}
\begin{wrapfigure}{R}{0.5\textwidth}
  \centering
  \includegraphics[width=0.5\textwidth]{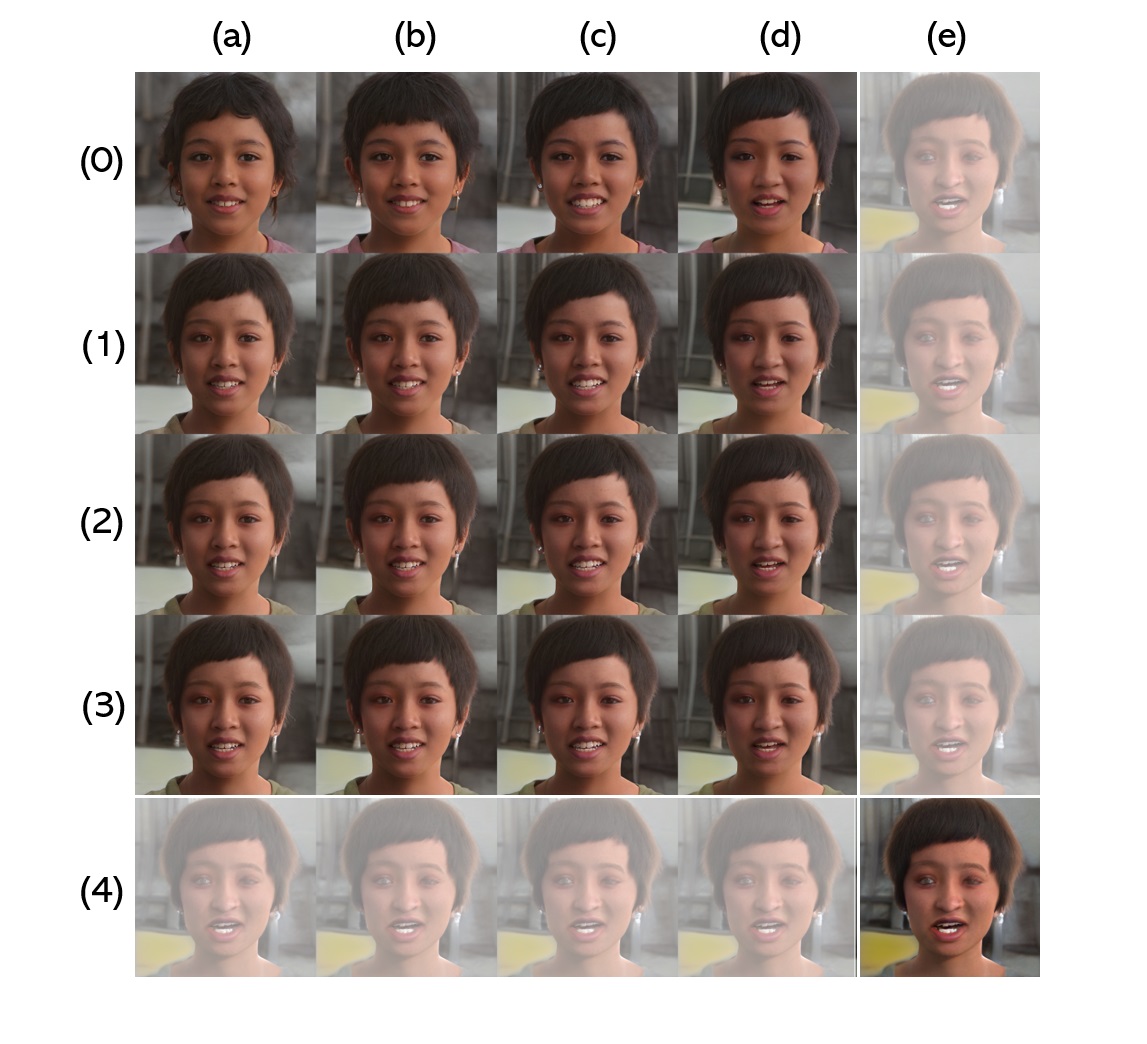}
  \vspace{-\intextsep}
  \caption{Interpolating beteen the output of step 2 ($a0$) and the output of the exact fit to the synthetics ($e4$). ($a-e$) represent the number of latent codes used for blending, and ($0-4$) the floating point weights for them.}
  \label{fig:interpolation_selection}
  \vspace{-\intextsep}
\end{wrapfigure}
To fit SG2 to the synthetic image $I^s$, we use the method of \cite{karras2019analyzing} with minor modifications based on empirical observation. We set the number of total steps to $1000$, the initial learning rate to $0.01$, and the initial additive noise to $0.01$. These changes are justified as we start from $w^n$ and so have a much-improved initialisation compared to the original algorithm. We also mask the loss using the same $I^a$ as above.

Having obtained a latent code $w^s$ that closely resembles the synthetic input image $I^s$, and a latent code that describes that apprximate nearest neighbour $I^n$, we can combine them in such a way that preserves the overall facial geometry of $I^s$ but has the fine detail of $I^n$. We use simple interpolation to do this, \textit{i.e.} the final latent code is obtained as:
\begin{equation}
w^{final} = w_s * \sqrt{\alpha} + w_n * \sqrt{1.0 - \alpha},
\end{equation}
where $w^{final}$ is a candidate for the final output of our method. We generate candidates by letting $\alpha$ retain the first $\{1, 3, 5, 7\}$ of the 18 latent codes with a floating point weight of $\{1.0, 0.9, 0.8, 0.7\}$ each. An example of the effect of this interpolation can be seen in Figure~\ref{fig:interpolation_selection}.

\subsection{Step 4: Sample Selection}
Having obtained a sequence of proposals, from step 1-4, we simply select the one that matches input most using the Structural Similarity (SSIM) \cite{MSSIM} metric at a resoluton of $368^2$ pixels, which we empirically found to give better qualitative results than the LPIPS distance. We hypothesise that this is due to the fact that perceptual losses prioritise texture over shape \cite{DBLP:journals/corr/abs-1811-12231}, and alignment of facial features is important for effective domain adaptation. We note that step 1-3 are run ten times with different random seeds to ensure that even difficult samples are matched with good solutions.

\section{Experiments}
We want to establish how realistic the images generated by our method are, and how well they preserve the semantics of the synthetic images, specifically head pose and facial features. To do so, we obtain a diverse set of $1000$ synthetic images, and process with them with our method, as well as two baselines.

We evaluate our algorithm quantitatively against the fitting method proposed in \cite{karras2019analyzing}, wich was designed to provide a latent embedding constrained to the domain of valid images in a pretrained SG2 model, and also operates at 1K resolution. This method is referred to as the \emph{StyleGAN2 Baseline}.

To assess how well we can match face pose and expression, we additionally compare facial landmark similarity as computed by OpenFace \cite{8373812}.

A qualitative comparison is made to the \emph{StyleGAN2 Baseline}, and CycleGAN \cite{CycleGAN2017}, with the latter trained on the entirety of our synthetic dataset. We conducted a user study to assess the perceived realism of our results compared to the \emph{StyleGAN2 Baseline} as well as the input images, and to provide an initial assessment of loss of semantics.

We make use of two variants of our results throughout this section. For \emph{Ours (only face)}, we replace the background and hair using the masks from the synthetic data by compositing them using a Laplacian Pyramid \cite{BURT1987671}. Because of the close alignment of our results with the input, this produces almost no visible artefacts. This allows us to ensure that the background does not impact the quantitive metrics., and to isolate just the appearance change of the face itself.

\subsection{Qualitative Experiments}
We train CycleGAN on FFHQ as well as a dataset of $12000$ synthetic images, using the default training parameters suggested by the authors. Despite having access to the synthetic training data, and using a much tighter crop, we found the results after 50 epochs unconvincing. Even at $128^2$, the images show artefacts, and lack texture detail. We show some results in Figure~\ref{fig:cycglegan_comparison}, and more in the supplementary material. Because of the overall quality of the results and because this method has access to the entire synthetic dataset during training, we do not include it in our user study.
\begin{figure}
  \centering
  \includegraphics[width=\textwidth]{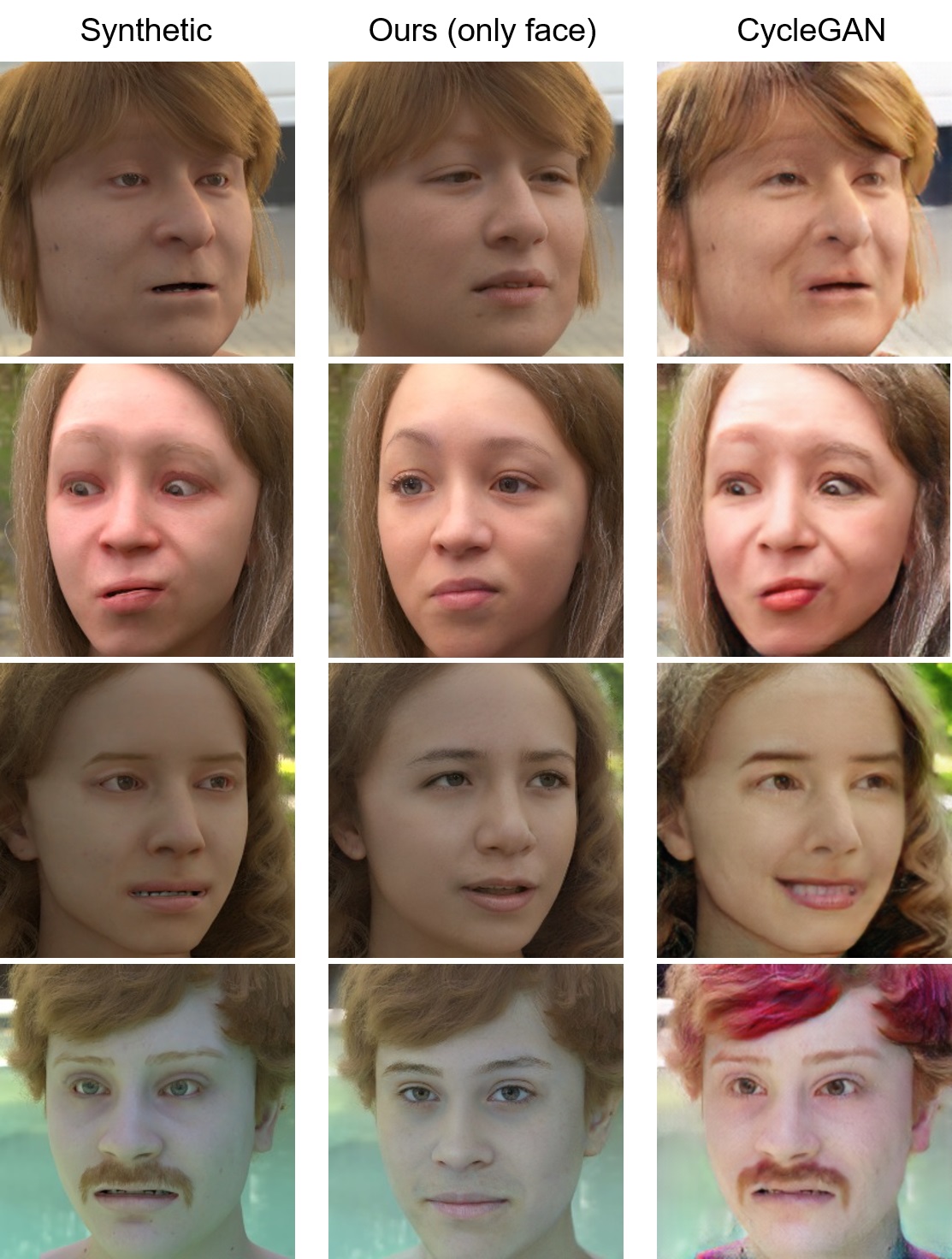}
  \caption{Representative comparison of results of our \emph{zero-shot} method vs CycleGAN trained on the whole synthetic dataset. Note how CycleGAN is unable to change the input images enough to make them look realistic. We suggest viewing this figure zoomed in.}
  \label{fig:cycglegan_comparison}
\end{figure}
Instead, we focus on the \emph{StyleGAN2 Baseline} for extensive evaluation.

We show each annotator three images: The synthetic input, the baseline result and our result, in random order. We ask if our result or the baseline is more photorealistic, and which image is the overall most realistic looking, \textit{i.e.} comparable to a real photograph.
Finally, we ask if the synthetic image and our result could be the same image of the same person. In this case, we let each annotator answer \{Definitely No, Slightly No, Slightly Yes, Definitely Yes\}.

From the annotation of 326 images, our results are considered more photoreal than the \emph{StyleGAN2 Baseline} in $94.48\%$ of cases. In $95.1\%$ of responses, our result was considered more realistic looking than the input or the baseline. 

In terms of whether the annotators thought the input and our result could be a photograph of the same person, the responses to the options \{Definitely No, Slightly No, Slightly Yes, Definitely Yes\} were selected $\{18.71, 19.1, 30.67, 31.6\}$ percent of the time. Despite the large gap in appearance, and the fact that our results are designed to alter aspects of the face like freckles which could be considered part of identity, roughly $60\%$ still believed our results sufficiently similar to pass as photograph of the same person at the same moment in time.

\begin{figure}
  \centering
  \includegraphics[width=\textwidth]{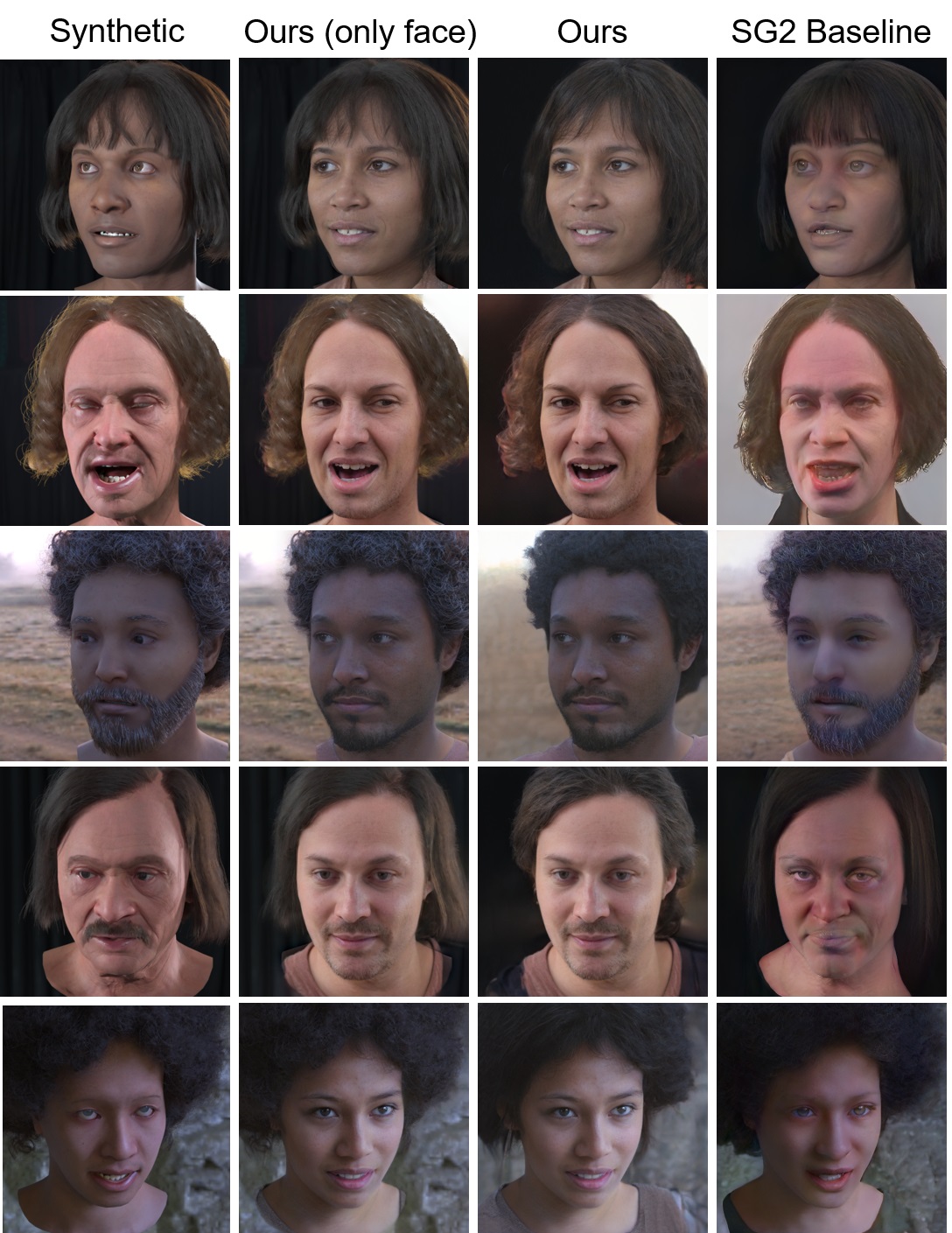}
  \caption{Representative comparison of results of our method vs the \emph{StyleGAN2 Baseline}. Both variants of our method are able to produce substantially more realistic samples with much greater detail. We suggest viewing this figure zoomed in.}
  \label{fig:ours_0}
\end{figure}

Figure~\ref{fig:ours_0} shows some of our results compared to the input synthetic images and the baseline.

\begin{wrapfigure}{R}{0.3\textwidth}
  \centering
  \includegraphics[width=0.3\textwidth]{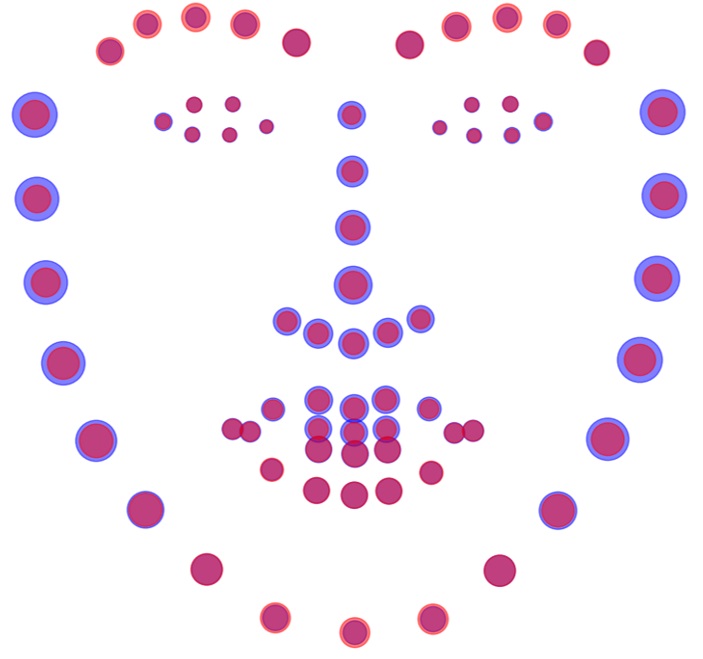}
  \vspace{-\intextsep}
  \caption{Standard deviation of the $L_1$ landmark error between the input and our results. \emph(scaled $20\times$ for figure). Blue/red indicate horizontal/vertical error.}
  \label{fig:lm_0}
  \vspace{-\intextsep}
\end{wrapfigure}
\subsection{Quantitative Experiments}
The preservation of important facial features is also assessed quantitatively by examing the alignment of $68$ landmarks \cite{8373812}. On our $1024^2$ images, the median absolute error in pixels is just $20.2$ horizontally, and $14.2$ vertically. We illustrate alignment errors per landmark in Figure~\ref{fig:lm_0}. The results indicate that the biggest errors occur on the boundary of the face near the ears, and that in the face region the eyebrows and lips have the highest degree of misalignment. Since FFHQ contains mostly smiling subjects, or images of people with their mouth closed, it is unsurpising that the diverse facial expressions from the synthetic data would show the greatest discrepancy in these features. We emphasize however, that these errors are less than two percent of the effective image resolution on average.

We also compute both the FID \cite{DBLP:journals/corr/HeuselRUNKH17} and IS \cite{DBLP:journals/corr/SalimansGZCRC16} metrics. The results are shown in Table~\ref{tab:large_small_crops} for the large and small crops used throughout this paper. Our method improves the FID signifcantly compared to the baseline, and slightly in case of the IS. This backs up the user study in terms of the perceptual plausibility of our results, but a larger number of samples would be beneficial for a conclusive result.

The FID difference between \emph{Ours} and \emph{Ours (only face)} shows that the background can significantly impact this metric, which is not reflected in human assessment.

\begin{figure}[]
  \centering
  \subfloat[IS and FID (to FFHQ) - large crop.]{
    \begin{tabular}[width=0.48\textwidth]{l|l|l|}
                     & IS & FID (to FFHQ) \\ \hline
  SG2 Baseline & 3.4965                    & 90.342                               \\ \hline
  Ours (only face)   & 3.3187                    & \textbf{78.728}                               \\ \hline
  Ours               & \textbf{3.653}                     & 81.06                                
  \end{tabular}
  }
  \centering
  \subfloat[IS and FID (to FFHQ) - tight crop.]{
\begin{tabular}[width=0.48\textwidth]{l|l|l|}
                     & IS           & FID (to FFHQ) \\ \hline
  SG2 Baseline & 3.398                     & 78.185                               \\ \hline
  Ours (only face)   & \textbf{3.464}                     & \textbf{70.731}                               \\ \hline
  Ours               & 3.435                     & 76.947                                 
  \end{tabular}
  }
  \centering
  \caption[]{We note that \emph{Ours (only face)} uses the same background and hair as the synthetic renders, while \emph{Ours} replaces the entire image with our fit. We hypothesise that the reason our results, which retain the background and hair from the synthetic renders, score better than those where the full image is optimised in terms of the FID, is because the background HDRs are visible in the renders. As these are natural images, we expect this to impact the metrics to some degree. Note that we resize the large crop to match CycleGAN resolution when calculating the IS.}
  \label{tab:large_small_crops}
\end{figure}

\section{Conclusions}
We have presented a novel zero-shot algorithm for improving the realism of non-photorealistic synthetic renders of human faces. The user study indicates that it produces images which look more photorealistic than the synthetic images themselves. It also shows that previous work on embedding images in the StyleGAN2 latent space produces results of inferior visual quality.

This result is reflected in quantitative terms in both the FID and IS metrics comparing our result to real images from FFHQ. 
CycleGAN, having access to a large dataset of synthetic images which our method never sees, and working on an inherently easier crop, is clearly not able to compare with our results qualitatively or quantitively as well.
A downside of our method is that it requires substantial processing time per image. We hypothesise that this could be amortised by training a model that predicts the StyleGan2 embeddings directly from synthetic images once a large enough dataset has been collected. We leave temporal consistency for processing animations as future work, and show more results as well as failure cases (for which the algorithm can simply be repeated with a different random seed) in the supplementary material.

\begin{figure}
  \centering
  \includegraphics[width=\textwidth]{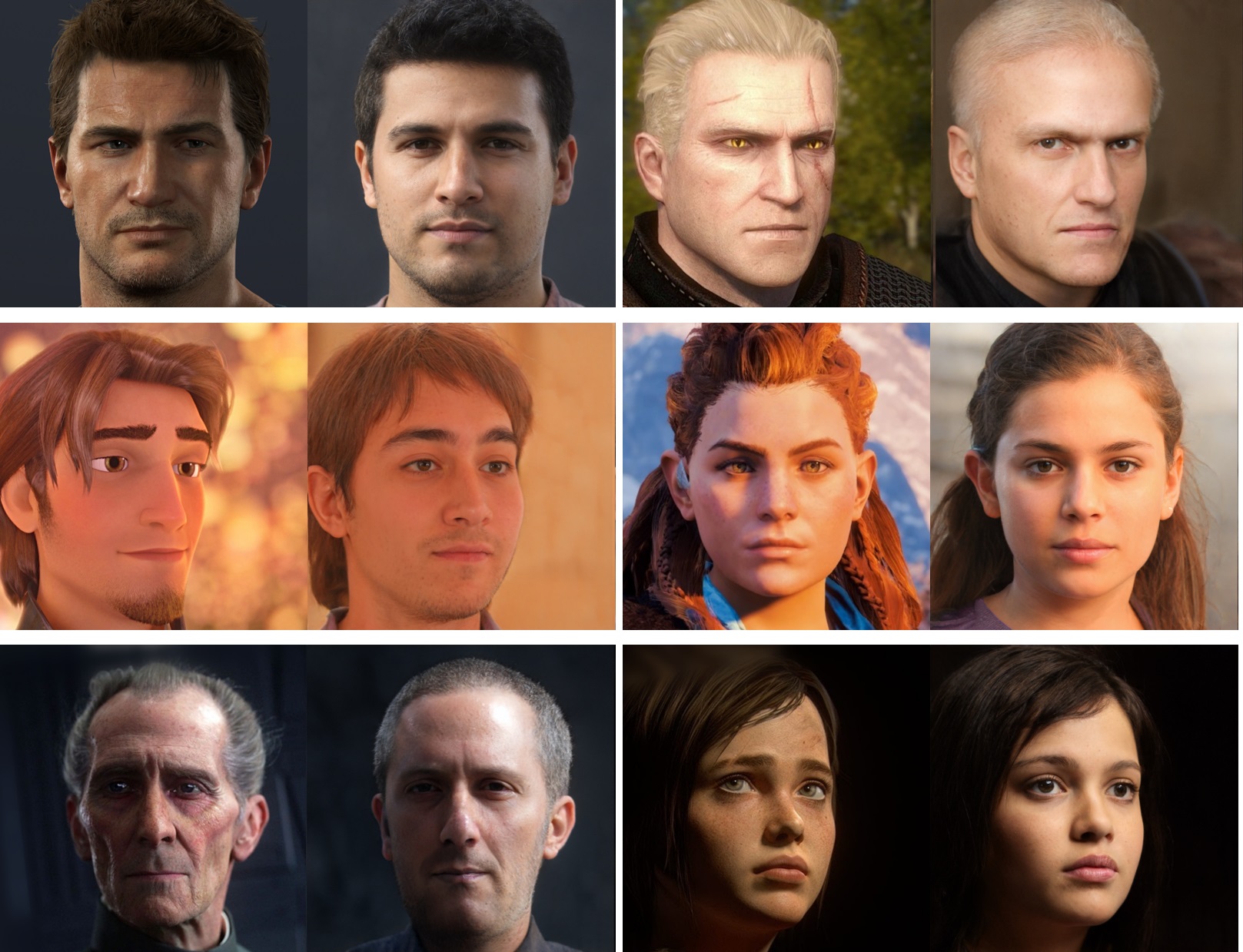}
  \caption{Our method applied to synthetic characters from popular culture. Left to right, row by row, these are: Nathan Drake from Uncharted, Geralt of Rivia from the Witcher, Flynn Rider from Tangled, Aloy from Horizon Zero Dawn, Grand Moff Tarkin from Rogue One, and Ellie from The Last of Us. We note that this is the output of only step 1 and 2 of our method. This indicates that we can find visually plausible nearest neighbours even with some exaggerated facial proportions.}
  \label{fig:ours_other_modalities}
\end{figure}

We would like to conclude by noting that our algorithm works across a wide range of synthetic styles (due to its zero-shot nature), and even with some non-photoreal images. Examples of this can be seen in Figure~\ref{fig:ours_other_modalities}.


\clearpage
%
%
\bibliographystyle{splncs04}
\bibliography{egbib}
\end{document}


\pagestyle{headings}
\mainmatter
\def\ECCVSubNumber{6059}  

\title{High Resolution Zero-Shot Domain Adaptation of Synthetically Rendered Face Images - \textbf{Supplementary Material}} 

\titlerunning{ECCV-20 submission ID \ECCVSubNumber} 
\authorrunning{ECCV-20 submission ID \ECCVSubNumber} 
\author{Anonymous ECCV submission}
\institute{Paper ID \ECCVSubNumber}

\maketitle

\section{Overview}
\begin{figure}
  \centering
  \includegraphics[width=\textwidth]{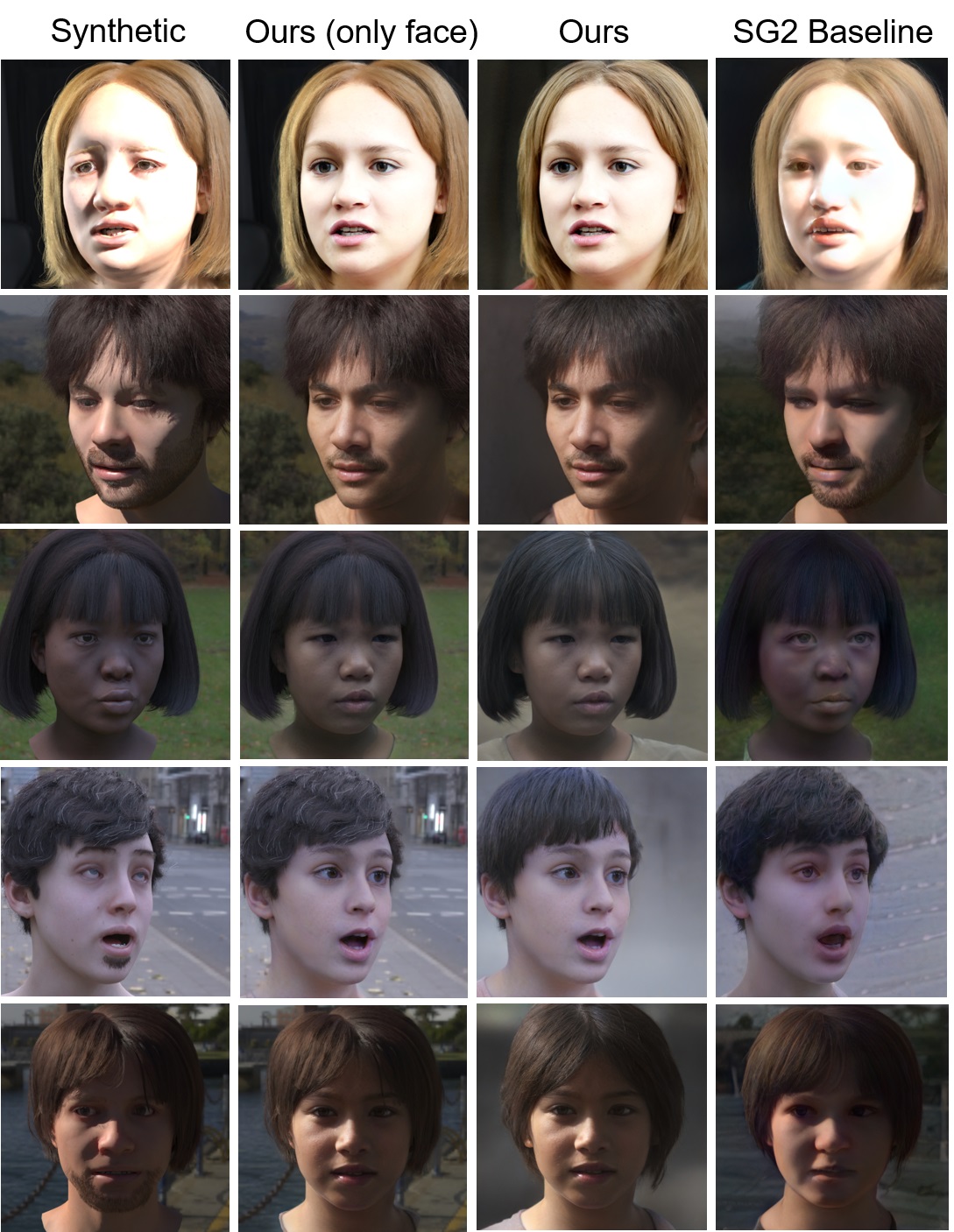}
  \caption{Yet more results of our method vs the \emph{StyleGAN2 Baseline}. We note that while our method preserves most of the synthetic image semantics, smaller beards, and eye gaze can be lost. Best viewed digitally, and in colour.}
  \label{fig:ours_2}
\end{figure}

We use the supplementary material to provide more detail on our method, specifically related to the control vectors and the CS-ANN algorithm. We also show more results (Figure~\ref{fig:ours_1} and Figure~\ref{fig:ours_2}), as well as some typical failure cases (Figure~\ref{fig:ours_failure}). Our method is entirely implemented in tensorflow \cite{tensorflow2015-whitepaper}.

\section{Control Vectors}
\begin{figure}
  \centering
  \includegraphics[width=\textwidth]{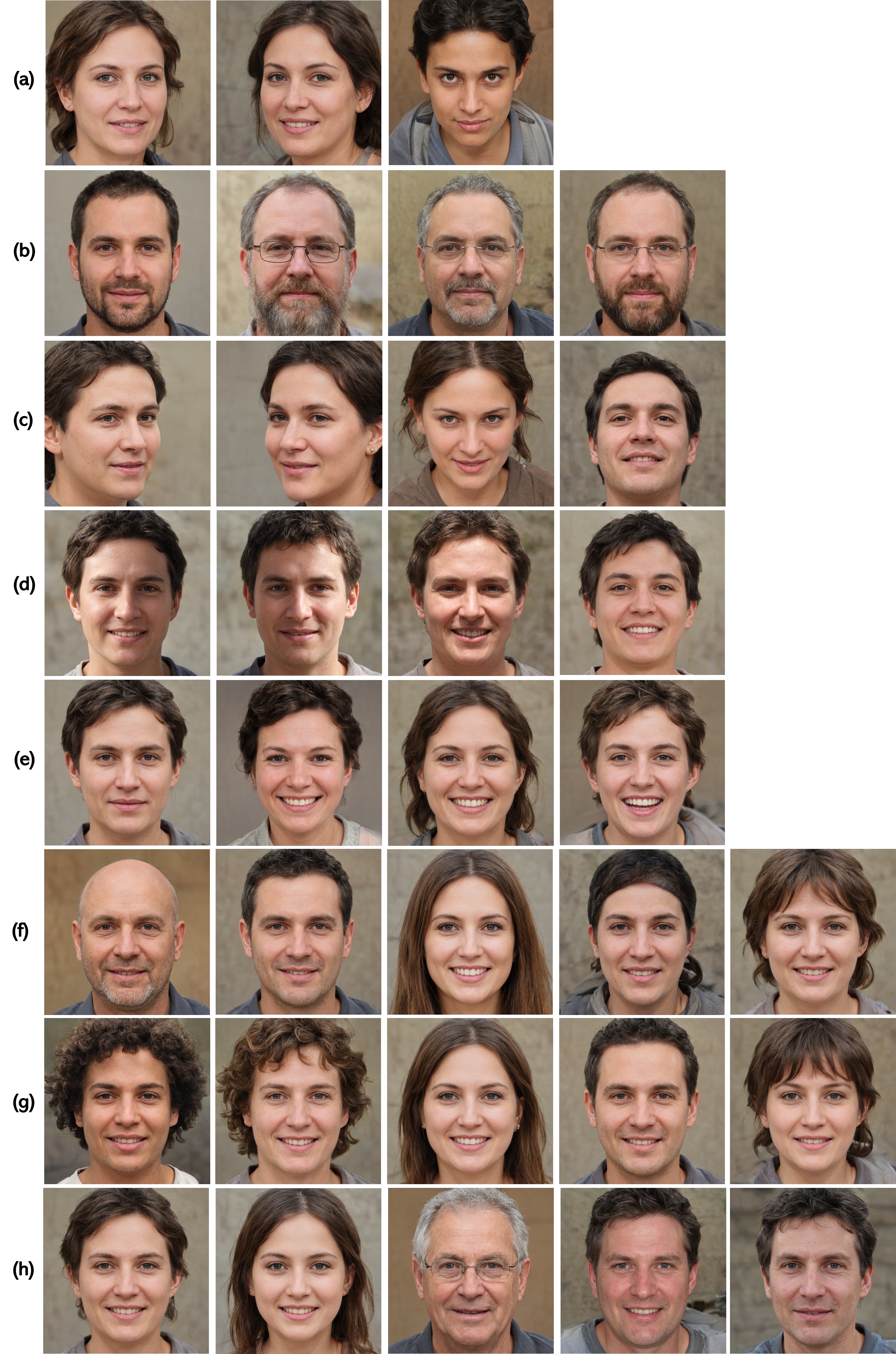}
  \caption{All control vectors, added to the 'mean' face of SG2. The categories are (a) Gaze Direction, (b) Beard Style, (c) Head Orientation, (d) Light Direction, (e) Degree of Mouth Open, (f) Hair Style, (g) Hair Type, (h) Skin Texture.}
  \label{fig:control_vectors_by_cat}
\end{figure}

The set of $33$ control vectors, $v_{control}$, is used in our method both during the sampling stage, and the CS-ANN algorithm. As with the centroids detailed in the main text, the vectors $v_{control}$ are selected purely empirically, based on what we perceived were problem cases for the fitting. We acknowledge that this could be automated in a more principled way in future. We show all control vectors used, added to the mean face of StyleGAN2 (SG2) \cite{karras2019analyzing} in Figure~\ref{fig:control_vectors_by_cat}. These vectors were obtained by annotating $2000$ samples from the prior of SG2 based on eight roughly defined categories: (a) Gaze Direction, (b) Beard Style, (c) Head Orientation, (d) Light Direction, (e) Degree of Mouth Open, (f) Hair Style, (g) Hair Type, (h) Skin Texture. Note that we do not claim that these are orthogonal. Interestingly, bias in the FFHQ dataset \cite{DBLP:journals/corr/abs-1812-04948} can be observed from $v_{control}$. For example, the gaze direction (a) is strongly correlated with the head orientation because subjects tend to look at the camera in photographs.

\section{Convex Set Approximate Nearest Neighbour Search}
After step 1 of the method returns $w^{s}$, we want to refine this latent code to map to an image better aligned to the input while staying in the space of plausible samples from $G$. The algorithm we propose and describe in the main text can be summarised as follows: 

\begin{algorithm}[H]
  \SetAlgoLined
  \KwResult{Refined latent codes, $w^n$}
   Initialise variables and fix random seeds\;
   \For{$i_{outer} = 0;\ i_{outer} < 96;\ i_{outer} += 1$}{
      Sample $512$ interpolation candidates\;
      Reset learnable parameters\;
      Set learning rate, $l = 0.01$\;
      \For{$i_{inner} = 0;\ i_{inner} < 20;\ i_{inner} += 1$}{
        Compute loss \& Update parameters\;
        \If{$i_{inner} \% 4 == 0$}{
          $l = l / 10.0$\;
        }
        Update current $w^n$ if better loss achieved\;
      }
    }
   \caption{Convex Set Approximate Nearest Neighbour Search}
   \label{alg:csanns}
  \end{algorithm}

While it would be possible to return the \emph{last} result of Algorithm~\ref{alg:csanns}, we found it slightly more effective to return the one that achieves the best loss across all iterations. To do this, we simply store the best loss and corresponding $w$ at any point in the outer or inner loop. If not using a learnable $\beta$ and the corresponding control vectors, we found that convergence was substantially slower. Thus, while it carries the danger of producing implausible results (as it no longer guarantees the result is a combination of samples), we allow $\beta$ to be learned. We note that clamping the range of $\beta$ during the optimisation appeared to be a sufficient constraint in practise.

As pointed out in the text, we allow different weights $\alpha$ for each of the $18$ SG2 inputs. We also experimented with a single weight per sample, using different alphas for each of the actual $18\times512$ inputs, as well as mixtures of these combinations. The former was substantially slower to converge while the latter was quick to produce implausible results.

For matching synthetic images with illumination substantially different from the dataset of real images, we also found it helpful to include a single learnable parameter (clamped to be in the range of $0.7-1.3$) to multiply the reproduced images with, thereby changing brightness by a simple linear transformation. We acknowledge that additional transformations such as histogram equalisation could be explored in future.

\section{Failure Cases}
\begin{figure}
  \centering
  \includegraphics[width=\textwidth]{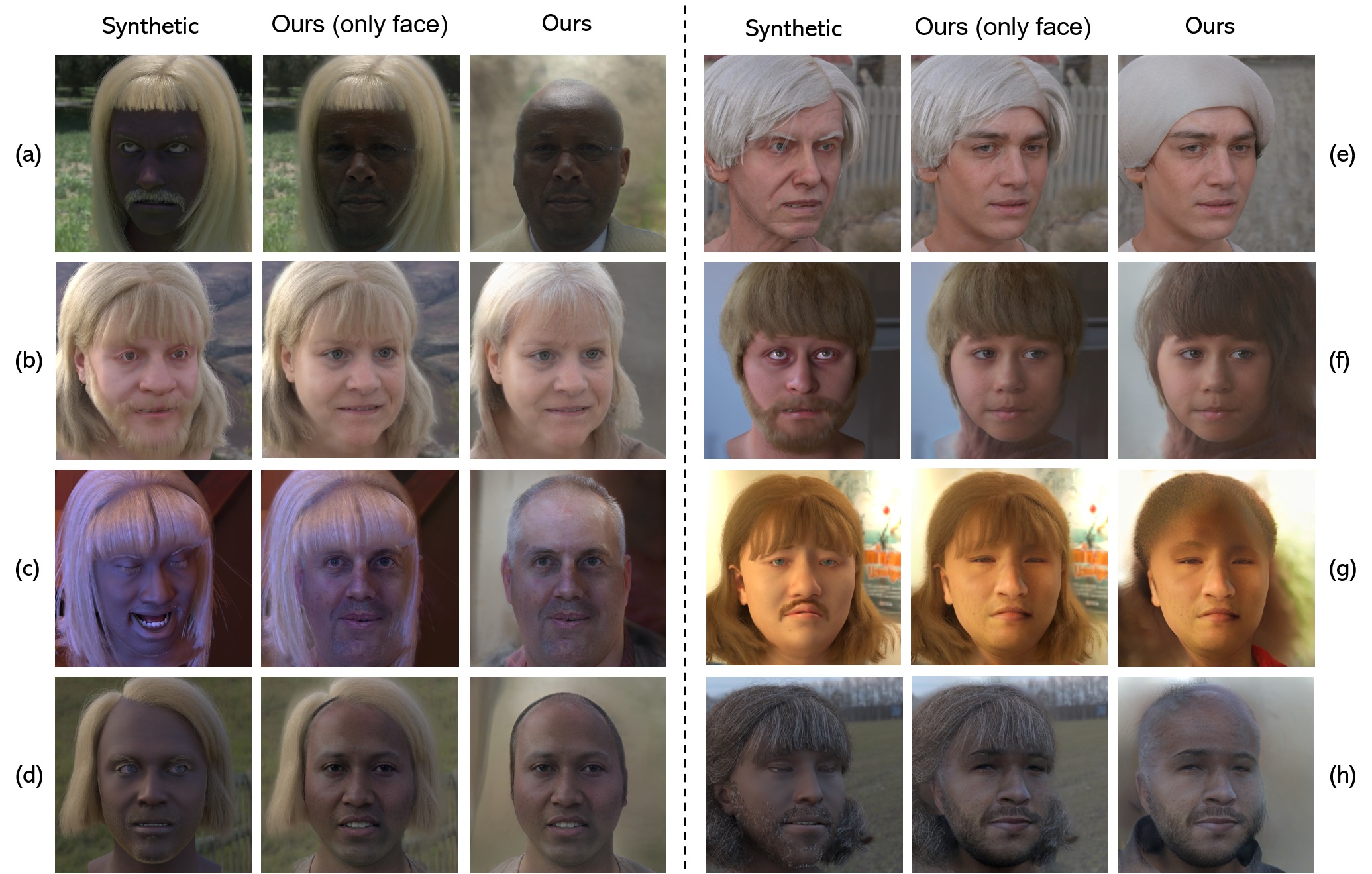}
  \caption{Example failure cases (persistent after 10 random starts of our method). Note how retaining the hair from synthetics in \emph{(ours (only face))} can help in such instances. It is of interest that the more unrealistic hair appearance in (e) is matched to a hat by our algorithm. The SG2 generator also appears to resist being steered with combinations unlikely to appear in FFHQ, such as (a) or (b).}
  \label{fig:ours_failure}
\end{figure}

We count any image for which our method fails to produce a relatively close match after $10$ runs with different random seeds as a failure case. As can be seen in Figure~\ref{fig:ours_failure}, we found very light hair, extreme poses / facial expressions and a lack of contrast between the face and hair challenging cases. We also observe that the further the synthetic image semantics are from the data space (i.e. FFHQ), the less likely our method is to produce satisfactory results. Cases (a) and (b) in Figure~\ref{fig:ours_failure}, combining beards and haircuts more typically associated with female faces, are examples of this. In such instances, the extension to our method that retains the hair from the synthetics can still produce plausible results as long as the facial shapes are approximately matched.

\begin{figure}
  \centering
  \includegraphics[width=\textwidth]{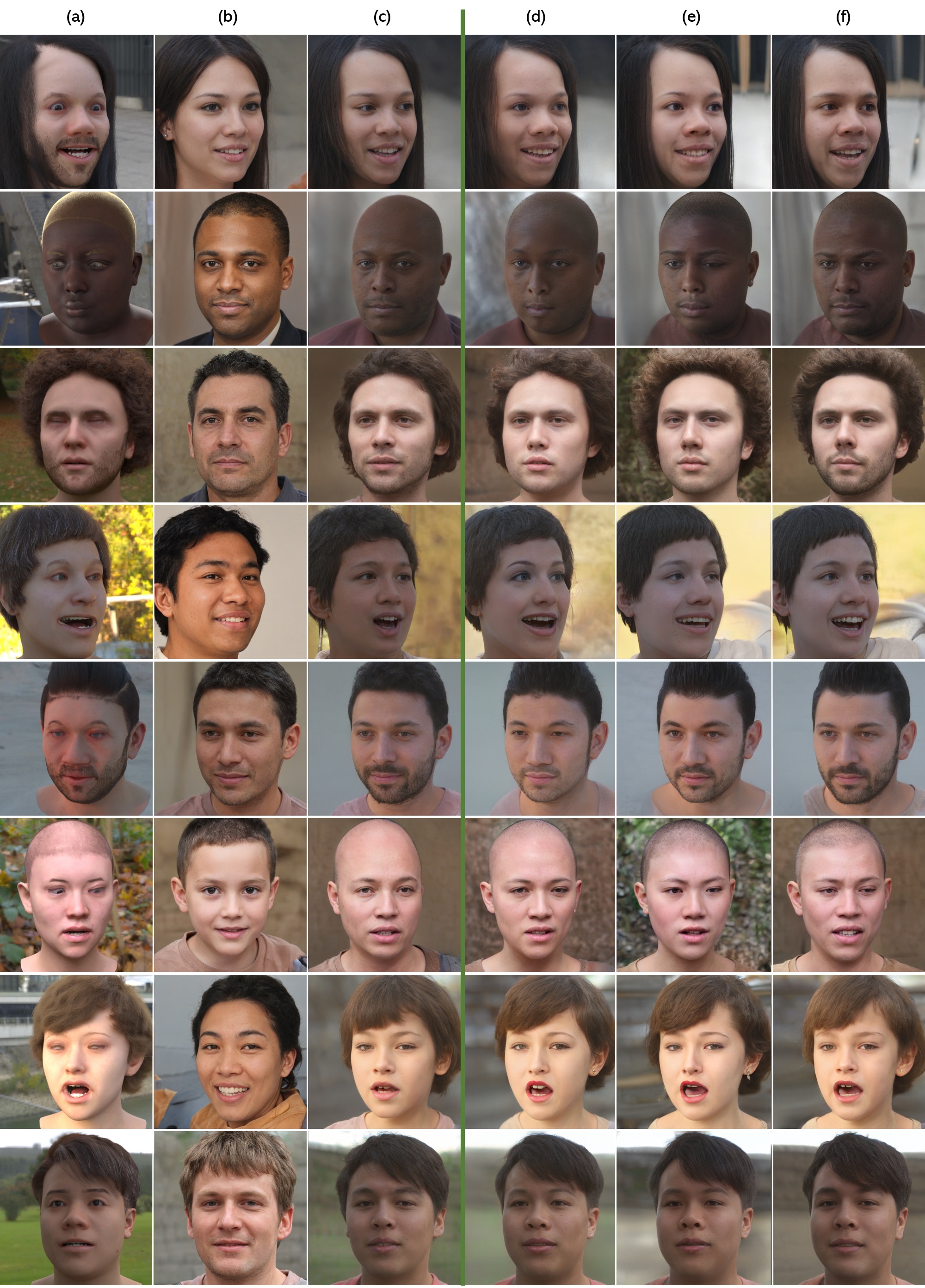}
  \caption{Extended illustration of the different steps of our method. (a) synthetic render, (b) output of the sampling in step 1, (c) the result of CS-ANN Search in step 2, and (d-f) results from step 3. Please refer back to the method section of the main text.}
  \label{fig:ours_1}
\end{figure}

\begin{figure}
  \centering
  \includegraphics[width=\textwidth]{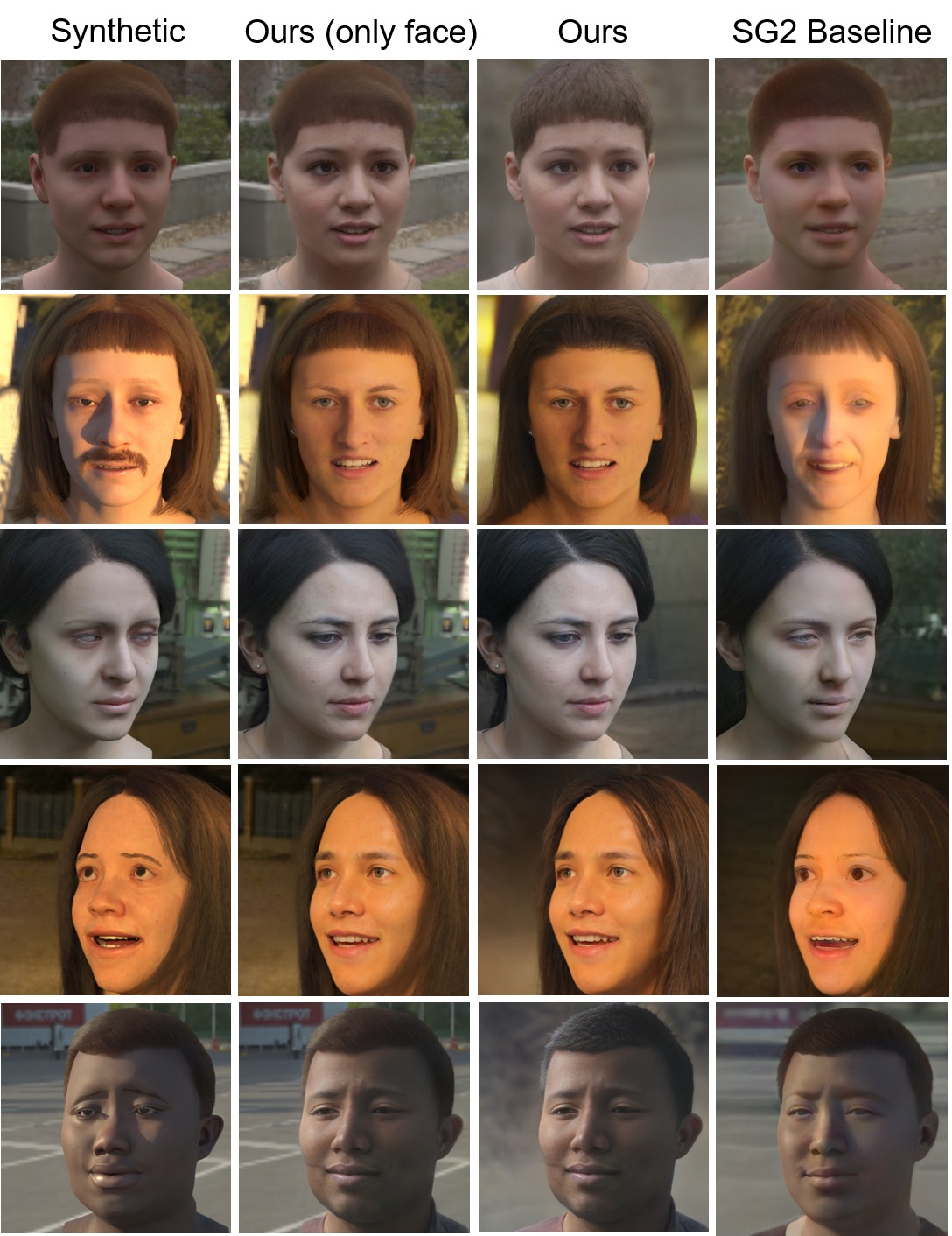}
  \caption{More results of our method vs the \emph{StyleGAN2 Baseline}. We note that while our method preserves most of the synthetic image semantics, smaller beards, and eye gaze can be lost. Best viewed digitally, as every image is at 1K resolution.}
  \label{fig:ours_1}
\end{figure}

\clearpage
%
%
\bibliographystyle{splncs04}
\bibliography{egbib}